\documentclass[sn-mathphys]{sn-jnl}

\usepackage{multicol}
\usepackage{multirow}
\usepackage{amsfonts} 
\usepackage{amssymb}
\usepackage{booktabs}
\usepackage{mathtools}
\usepackage{tabularx}
\usepackage{graphicx}
\usepackage{adjustbox}

\usepackage{colortbl}  
\usepackage{xcolor}
\usepackage{amsmath}
\usepackage{amsthm}
\usepackage{mathrsfs}
\usepackage{pifont}
\usepackage{MnSymbol}
\usepackage{balance}
\usepackage{algorithm}
\usepackage{algpseudocode}

\jyear{2021}%

\theoremstyle{thmstyleone}%
\theoremstyle{thmstyletwo}%

\theoremstyle{thmstylethree}%

\begin{document}

\title[StegaFFD: Privacy-Preserving Face Forgery Detection]{{StegaFFD: Privacy-Preserving Face Forgery Detection via Fine-Grained Steganographic Domain Lifting}}


\author[1,2]{\fnm{Guoqing} \sur{Ma}}
\equalcont{These authors contributed equally to this work.}

\author[3]{\fnm{Xun} \sur{Lin}}
\equalcont{These authors contributed equally to this work.}


\author[4]{\fnm{Hui} \sur{Ma}}

\author[1,4]{\fnm{Ajian} \sur{Liu}}

\author[3]{\fnm{Yizhong} \sur{Liu}}

\author[3]{\fnm{Wenzhong} \sur{Tang}}

\author[1,2]{\fnm{Shan} \sur{Yu}}

\author[5]{\fnm{Chenqi} \sur{Kong}}

\author*[5]{\fnm{Yi} \sur{Yu}}\email{yu.yi@ntu.edu.sg}

\affil[1]{\orgdiv{Institute of Automation}, \orgname{Chinese Academy of Sciences}, \orgaddress{\city{Beijing} \postcode{100190},  \country{China}}}

\affil[2]{\orgdiv{School of Future Technology}, \orgname{University of Chinese Academy of Sciences}, \orgaddress{\city{Beijing} \postcode{100049},  \country{China}}}

\affil[3]{ \orgname{Beihang University}, \orgaddress{\city{Beijing} \postcode{100191},  \country{China}}}

\affil[4]{ \orgname{Macau University of Science and Technology}, \orgaddress{\city{Macau SAR} \postcode{999078},  \country{China}}}

\affil[5]{\orgdiv{ROSE Lab, School of EEE}, \orgname{Nanyang Technological University}, \orgaddress{\city{Nanyang Avenue} \postcode{639798},  \country{Singapore}}}


\abstract{
Most existing Face Forgery Detection (FFD) models assume access to raw face images. In practice, under a client-server framework, private facial data may be intercepted during transmission or leaked by untrusted servers. Previous privacy protection approaches, such as anonymization, encryption, or distortion, partly mitigate leakage but often introduce severe semantic distortion, making images appear obviously protected. 
This alerts attackers, provoking more aggressive strategies and turning the process into a cat-and-mouse game. 
Moreover, these methods heavily manipulate image contents, introducing degradation or artifacts that may confuse FFD models, which rely on extremely subtle forgery traces.
Inspired by advances in image steganography, which enable high-fidelity hiding and recovery, we propose a \textbf{Stega}nography-based \textbf{F}ace \textbf{F}orgery \textbf{D}etection framework (StegaFFD) to protect privacy without raising suspicion. 
StegaFFD hides facial images within natural cover images and directly conducts forgery detection in the steganographic domain. 
However, the hidden forgery-specific features are extremely subtle and interfered with by cover semantics, posing significant challenges. 
To address this, we propose Low-Frequency-Aware Decomposition (LFAD) and Spatial-Frequency Differential Attention (SFDA), which suppress interference from low-frequency cover semantics and enhance hidden facial feature perception. Furthermore, we introduce Steganographic Domain Alignment (SDA) to align the representations of hidden faces with those of their raw counterparts, enhancing the model's ability to perceive subtle facial cues in the steganographic domain.
Extensive experiments on seven FFD datasets demonstrate that StegaFFD achieves strong imperceptibility, avoids raising attackers' suspicion, and better preserves FFD accuracy compared to existing facial privacy protection methods.
}


\maketitle

\section{Introduction}

\noindent 
Compared with fingerprints, irises, and voiceprints, the face image is more intuitive, has richer information, a lower acquisition cost, and can be used in a non-contact manner. 
Therefore, it has been extensively used in identity authentication and identification services~\cite{thies2016face2face,xun2025tpami,xun2024cvpr,rizhao2025tpami}. 
However, face images are easy to forge. 
In recent years, with the rise of face forgery technology and tools like FaceApp, FaceSwap, and DeepFaceLab~\cite{li2021identity}, even non-professionals can manipulate face images, leading to widespread fake faces online. 
This poses significant security risks in areas such as justice, politics, reputation, and criminal investigations~\cite{xun2024patterns,lin2023image,cdsnet}, necessitating face forgery detection (FFD) techniques~\cite{law2024tifs,law2025tcsvt,kcq2022tifs,moe-kcq,gmdf}. 
In recent years, significant attention has been paid to face forgery detection, with models based on physiological signal analysis~\cite{yang2019exposing}, intrinsic features~\cite{nataraj2019detecting}, and deep learning~\cite{lin2023deepfake, ma2022improving, ma2023boosting}. 
Deep learning-based models, in particular, have shown promise in effectively distinguishing real faces from face-forgery ones.

\begin{figure*}
\centering
\includegraphics[width=1.02\textwidth]{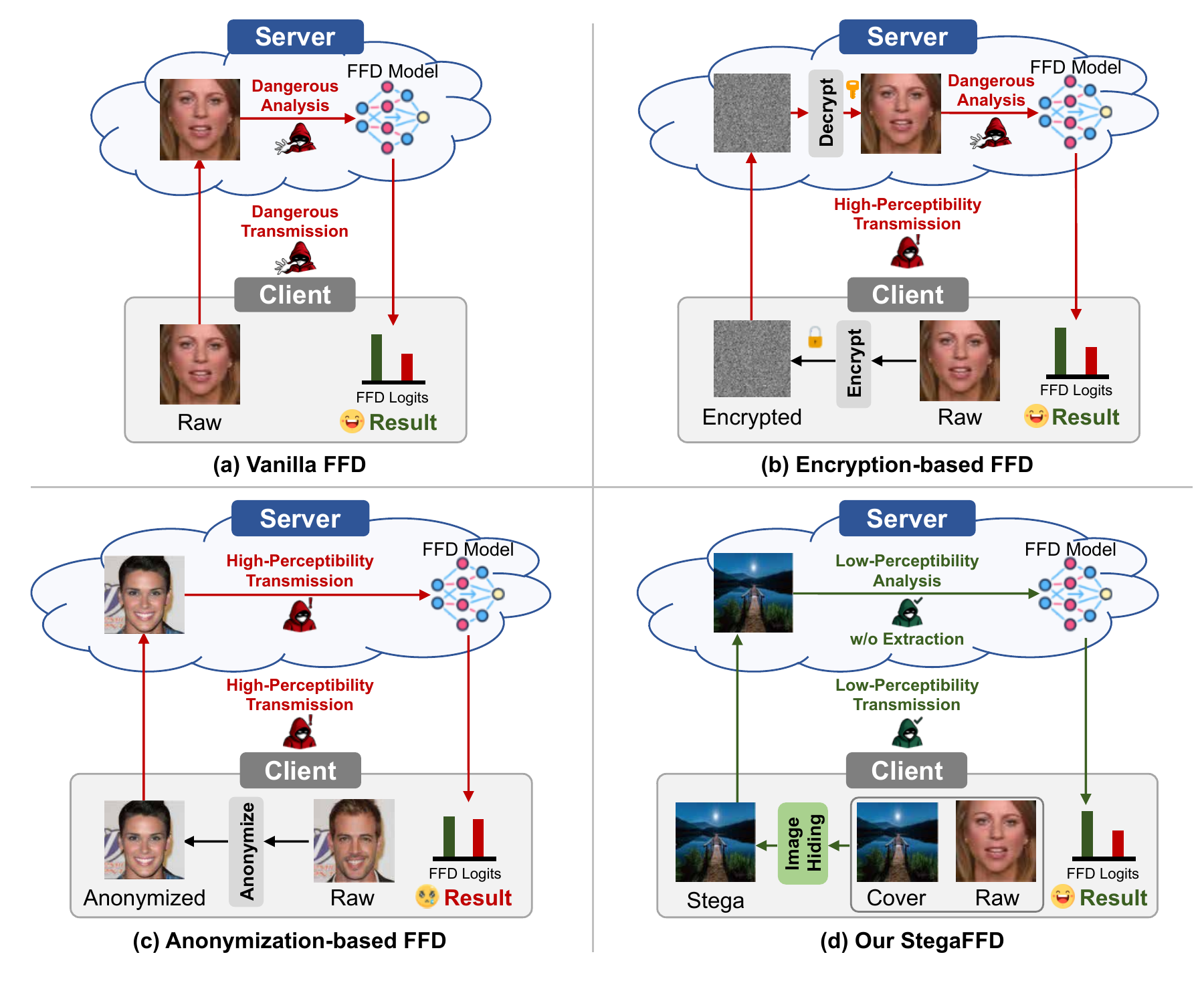}
  \caption{Different client-server frameworks for FFD, including (a) vanilla, (b) encryption-based, (c) anonymization-based, and (d) the proposed StegaFFD framework. Our StegaFFD makes the attacker fail to notice the existence of facial images.}
  \label{fig:intro}
\end{figure*}

%
Despite the impressive performance of deep neural networks, there are increasing concerns about security issues~\cite{meng2024semantic,yu2022towards,Yu_2023_CVPR,yu2024purify,yu2024robust,yu2025backdoor,yu2025towards,yu2025mtlue,wu2025temporal} associated with artificial intelligence, especially with the client-server architecture. Among these security issues, the protection of users' privacy remains a severe challenge~\cite{luo2018privacyprotector}. From a legal and ethical perspective, people's facial images must be rigorously protected, ensuring the security and privacy of these images during their storage, transmission, and analysis processes~\cite{kaissis2020secure} in client-server scenes. 
Institutions managing facial data have adopted methods such as data anonymization or pseudonymization~\cite{bagai2017measuring} to mitigate the risk of data leakage. 
However, real-world scenarios indicate that these methods are not robust against re-identification attacks~\cite{kaissis2021end}. 

As illustrated in Fig.~\ref{fig:intro}~(a), there is a risk of leakage, especially during the transmission of images from a client to a server~\cite{avudaiappan2018medical}. 
Although techniques like non-homomorphic image encryption can prevent image leakage during transmission, as shown in Fig.~\ref{fig:intro}~(b), they are not robust to attacks from malicious servers~\cite{deform1}. 
Once an attacker controls the server through system vulnerabilities or embedded malicious backdoors (as a third-party server provider), the decrypted images on the server-side are also at risk of being captured. 
Face anonymization methodologies~\cite{li2021learning} induce synthetic characteristics in genuine images, consequently degrading the detection efficacy of face forgery detection algorithms, as shown in Fig.~\ref{fig:intro}~(c). 
We position StegaFFD as a covert client–server FFD framework that avoids revealing facial content during transmission and requires no decryption on the server, as shown in Fig.~\ref{fig:intro}~(d). {It is noteworthy that the server returns to the client the "logit indicating whether the face image is forged." This process does not pose a privacy leakage issue. Since attackers are likely to assume that our input image is a normal image without faces, it is difficult for them to discern the meaning of this logit.}

%
Currently, most image privacy enhancement technologies focus on the training phase, for instance, by enabling different institutions to collaboratively train models without the direct exchange of image data through federated learning~\cite{guan2024federated, liu2024bfl}. These methods effectively enhance the privacy protection of training data. Some studies focus on protecting images after deployment through distortion-based~\cite{compress, deform1, deform2} methods to modify detailed information of the original images. Although these methods somewhat protect image privacy, attackers can easily notice that the transmitted images are facial images. To further enhance security, recent works propose encoding-based~\cite{encode2, encode1} and homomorphic-encryption-based~\cite{encrypt2, encrypt1} methods to protect facial images. However, these methods significantly alter the distribution of the images, easily alerting attackers to the fact that the images are protected by specific techniques, thus falling into a cat-and-mouse game of encryption and decryption. Among these methods, methods based on homomorphic encryption are time-consuming, making them impractical for many real-world applications, as shown in Table~\ref{tab:Comparison}. 

%
Recent progress in deep image hiding (DIH) networks~\cite{deepmih, hinet, pusnet}, which are proposed to conceal a secret image within a cover image for covert image transmission, has inspired us to propose a novel framework to perform Face Steganographic Domain Analysis (StegaFFD) based on DIH. As shown in Fig.~\ref{fig:intro}~(d), StegaFFD hides a facial image within a natural photo, allowing for direct FFD in the steganographic domain without the image extraction and decryption on the server-side. 
However, conducting FFD directly on stego images is challenging. The facial-specific features crucial for FFD are intrinsically weak, and their signal becomes even fainter in the steganographic domain. Meanwhile, the dominant information from the cover image acts as pervasive noise, effectively drowning out these critical but feeble cues. 


Stego images carry most scene semantics in low-frequency bands, whereas DIH typically embeds secrets in relatively higher-frequency details to remain imperceptible~\cite{hinet,hidemia,xin2023iccv}. 
We thus estimate cover semantics via LFAD (frequency-aware Low-Frequency-Aware Decomposition) and suppress them via SFDA (Spatial-Frequency Differential Attention), yielding secret-focused features for FFD.
These two modules are integrated with StegaFFD's server-side facial image analysis network. They can perform frequency-adaptive analysis on stego images in the steganographic domain (separating semantic information and secret information~\cite{chen2024frequency}). 
Specifically, considering that most of the semantic information in the cover belongs to the low-frequency bands, we design LFAD inspired by~\cite{chen2024frequency2} with low-pass filters to extract the cover image information. 
To prevent blurring at boundaries, inspired by~\cite{yang2021simam}, it predicts spatially variant low-pass filters for the feature instead of using the fixed kernel in the conventional interpolation~\cite{chen2024frequency2}. 
Under this operation, LFAD extracts the cover information from concealed facial images, which aids the subsequent SFDA in removing the cover information and extracting the secret information.
Next, we propose Spatial-Frequency Differential Attention (SFDA) to enable a global perception of concealed facial images across spatial and spectral dimensions. Through the Differential Transformer, we remove semantic information in the cover images, preventing cover-specific features in the low-frequency band from hindering the extraction of facial related features in the high-frequency bands while ensure global perception. 
%

Moreover, to enhance the practicality of the StegaFFD system and further ensure the accuracy of FFD, motivated by~\cite{li2023object}, we propose the Steganographic Domain Alignment (SDA) with an auxiliary network to enable the detector to retain more facial and forged features. 
The training and detection processes are decoupled, meaning that SDA is only employed during training, thereby eliminating the risk of data leakage during detection. During training, SDA extracts clean facial-related features and distills them into the detector network, further enhancing the accuracy of FFD.

Our contributions can be summarized as follows:
\begin{itemize}
    \item We propose a privacy-enhancing client-server FFD framework, namely StegaFFD, to covertly analyze facial images without raising the suspicions of attackers. 
    \item We design LFAD and SFDA for frequency-aware stego images to ensure that StegaFFD can accurately perform FFD in the steganographic domain without interference from conspicuous information in the covers. 
    \item We develop SDA, which, through the proposed steganographic domain alignment method, further improves the accuracy of the FFD network enhancing the practicality of the StegaFFD.
    \item Extensive experiments on seven datasets demonstrate the effectiveness and imperceptibility of StegaFFD.
\end{itemize}

\section{Related works}
\subsection{Facial Privacy Protection} 
\noindent To mitigate privacy risks in face recognition, privacy-preserving face recognition (PPFR) methods operate in protected domains. These methods fall into two categories: (1) Cryptography-Based Methods, which perform recognition on encrypted images using techniques like garbled circuits~\cite{sadeghi2009efficient}, homomorphic encryption~\cite{huang2020instahide}, and secure multiparty computation~\cite{ma2019lightweight}. They maintain accuracy but suffer from high computational costs and lack visualization utility, easily alerting attackers. (2) Transformation-Based Methods, which apply frequency-domain learning~\cite{wang2022privacy}, generative models~\cite{li2019privacy}, and differential privacy~\cite{ergun2014privacy}. 
%

Identifiable Face Anonymization (IDFA) aims to anonymize users' identities while preserving facial characteristics. Existing methods include image obfuscation~\cite{winkler2010trustcam} and face synthesis~\cite{hukkelaas2019deepprivacy}. Some methods leverage deep models to recognize privacy-protected faces~\cite{zhao2017dual}. Li et al.~\cite{li2021learning, li2021identity} introduced identity-aware anonymization using GANs. Yuan et al.~\cite{yuan2022pro} and Zhang et al.~\cite{zhang2023rapp} proposed steganography-based methods. Wang et al.~\cite{wang2023identifiable} developed a virtual identity transformation. However, since IDFA modifies images in the original image domain, it transforms non-forged images into forged-like data. As a result, the performance of face forgery detection based on IDFA images is compromised.
While PPFR and IDFA focus on recognition under protection or identity masking, they alter image distribution or reveal processing traces, making them ill-suited for FFD, where fidelity to forgery cues is critical.

\subsection{Deep Image Hiding}

\noindent Image hiding aims to covertly conceal a secret image within a cover image while allowing for accurate extraction of the hidden content. The first deep learning-based image hiding (DIH) networks were introduced by Baluja~\cite{deepstega}, and Hayes and Danezis~\cite{advstega}, utilizing an encoder-decoder architecture. To improve embedding efficiency,~\cite{waveletstega} incorporated U-Net and discrete wavelet transformation (DWT)~\cite{dwt}, enhancing the reversibility of hidden images. The development of invertible neural networks (INNs) in various image-to-image tasks~\cite{rescale,sp} has further advanced image hiding techniques. 
An INN-based framework is introduced to formulate image embedding and extraction as forward and inverse affine transformations~\cite{large}. To achieve higher reversibility and imperceptibility, Deng et al.~\cite{hinet, deepmih} proposed an INN-based method where both the cover and secret images undergo wavelet transformation. This technique embeds the secret image into the high-frequency components of the cover image, minimizing visible distortions while ensuring accurate recovery.

\subsection{Face Forgery Detection}


%

\noindent Face forgery detection initially relied on image processing methods (e.g., copy-move, splicing, deletion) to extract inherent features and train machine learning classifiers, such as Speeded-Up Robust Features (SURF)~\cite{zhang2017automated} and Photo Response Non-Uniformity (PRNU)~\cite{koopman2018detection}. However, these methods often fail against deep learning-generated forgeries (e.g., GAN, VAE). Recent approaches focus on deep networks, including plain deep detector~\cite{afchar2018mesonet, rossler2019faceforensics++, zhang2025efficient}, spatial deep detector~\cite{nguyen2019multi, nguyen2019capsule}, and frequency deep detector~\cite{qian2020thinking, liu2021spatial, durall2020watch}. 1) Plain detector employs CNNs to directly distinguish face forgery content from authentic data. Numerous CNN-based binary classifiers have been proposed, e.g., MesoNet~\cite{afchar2018mesonet} and Xception~\cite{rossler2019faceforensics++}. 2) Spatial detector delves deeper into specific representations, such as forgery region location~\cite{nguyen2019multi}, capsule network~\cite{nguyen2019capsule}, disentanglement learning~\cite{yan2023ucf, ma2025clustering, ma2025mitigating}, image reconstruction~\cite{cao2022end}, erasing technology~\cite{wang2021representative}, etc. Besides, some other methods specifically focus on the detection of blending artifacts~\cite{li2018exposing}, generating forged images during training in a self-supervised manner to boost detector generalization. 3) Frequency detector addresses this limitation by focusing on the frequency domain for forgery detection~\cite{durall2020watch}. SPSL~\cite{liu2021spatial} and SRM~\cite{luo2021generalizing} are other examples of frequency detectors that utilize phase spectrum analysis and high-frequency noises, respectively. Qian et al.~\cite{qian2020thinking} propose the use of learnable filters for adaptive mining of frequency forgery clues using frequency-aware image decomposition.
%

{\subsection{Privacy Protection in FFD}
Previous privacy protection approaches for FFD, such as anonymization, distortion, and encryption, partly mitigate leakage but often introduce severe semantic distortion, making images appear obviously protected. 
Face anonymization methodologies~\cite{li2021learning} induce synthetic characteristics in genuine images, consequently degrading the detection efficacy of face forgery detection algorithms. 
Some studies focus on protecting images after deployment through distortion-based~\cite{wen2015face} methods to modify detailed information of the original images. Although these methods somewhat protect image privacy, attackers can easily notice that the transmitted images are facial images. 
To further enhance security, some approaches utilize entropy-based methods~\cite{chen2023privacy}. 
However, these methods significantly alter the distribution of the images, easily alerting attackers to the fact that the images are protected by specific techniques, thus falling into a cat-and-mouse game of encryption and decryption. Among these methods, methods based on homomorphic encryption are time-consuming, making them impractical for many real-world applications. }

\begin{table*}[t]

\caption{Comparison between StegaFFD and other {prominent} privacy-protection approaches ensuring the security and privacy of facial images during their training, transmission, and analysis processes. The plaintext or encrypted forms of information may raise alerts among malicious actors. }

\setlength{\tabcolsep}{4pt} 

\resizebox{1.\textwidth}{!}{
\begin{tabular}{l|ccc}

\toprule
{\textbf{Protection Method}} & {\textbf{Protection Target}}&  {\textbf{Alert Raising}} & {\textbf{Acc. Retention}} \\ 
\midrule
No Protection & --- & \textcolor{red!80!black}{\ding{51}} (Facial Image) & \textcolor{green!80!black}{\ding{51}}  \\
\midrule
Federated Learning~\cite{guan2024federated, liu2024bfl} & Training Data & --- & \textcolor{green!80!black}{\ding{51}}  \\
\midrule
Differential Privacy~\cite{abadi2016deep} & Training Data & --- & \textcolor{green!80!black}{\ding{51}}  \\
\midrule
Encryption~\cite{deform1} & Testing Data & \textcolor{red!80!black}{\ding{51}} (Noise)  & \textcolor{green!80!black}{\ding{51}} \\
\midrule
Distortion~\cite{compress} & Testing Data & \textcolor{red!80!black}{\ding{51}} (Distorted Facial Image)  & \textcolor{red!80!black}{\ding{55}}  \\
\midrule
Anonymization~\cite{li2021learning} & Testing Data & \textcolor{red!80!black}{\ding{51}} (Facial Image)  & \textcolor{red!80!black}{\ding{55}}  \\
\midrule
StegaFFD (Ours) & Testing Data & \textcolor{green!80!black}{\ding{55}} (Natural Image) & \textcolor{green!80!black}{\ding{51}}   \\
\bottomrule
\end{tabular}
}

\label{tab:Comparison}

\end{table*}

\section{Methodology}

\subsection{Overall Framework}
\begin{figure*}
    \centering
    \includegraphics[width=1.0\textwidth]{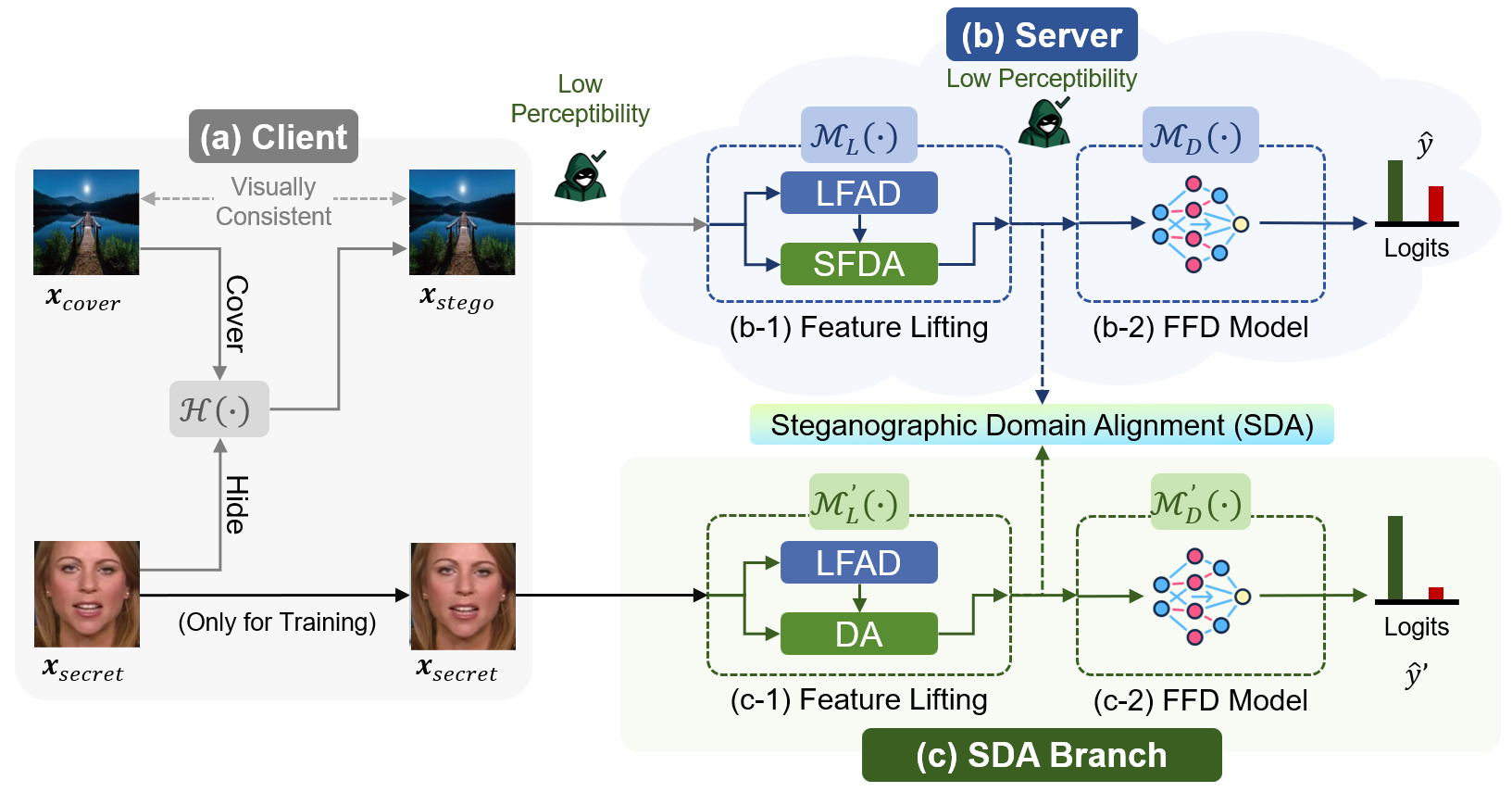}
    \caption{Illustration of the detailed structure of the proposed StegaFFD. (a) Client-side image hiding network $\mathcal H(\cdot)$ hides facial secret images into natural cover images for privacy-preserving transmission. (b) Server-side Detector analyzes stego images directly in the steganographic domain to lift facial features by $\mathcal{M}_L$ for forgery detection. (c) SDA branch enhances detector training by aligning steganographic-domain features with raw facial features lifted by  $\mathcal{M}'_L$, improving forgery detection accuracy. This branch operates exclusively during training and is omitted during deployment. } 
    \label{fig:main}
\end{figure*}

Our StegaFFD consists of an image hiding network $\mathcal H(\cdot,\cdot)$, a facial image analysis network $\mathcal M(\cdot)$, and a steganographic domain alignment network $\mathcal M'(\cdot)$, as illustrated in Fig.~\ref{fig:main}.
The image hiding network $\mathcal H(\cdot,\cdot)$ is deployed on the client side, while the facial image analysis network $\mathcal M(\cdot)$ runs on the server.
The steganographic domain alignment network $\mathcal{M'}(\cdot)$ assists $\mathcal M$ in learning facial-relevant features in the steganographic domain and is used only during training.
Let $\boldsymbol x_{secret}$ denote the raw facial image captured by user devices (e.g., smartphones or laptops), and let $\boldsymbol x_{cover}$ denote a natural image used as the cover. The hiding process
$\boldsymbol x_{stego}=\mathcal H(\boldsymbol x_{secret},\boldsymbol x_{cover})$
produces a stego image in which $\boldsymbol x_{secret}$ is embedded into $\boldsymbol x_{cover}$. After concealment, $\boldsymbol x_{stego}$ is transmitted from the client to the server and fed directly to the facial image analysis network, yielding the final prediction
$\boldsymbol{\hat{y}}=\mathcal M(\boldsymbol x_{stego})$.
The facial image analysis network analyzes the concealed facial content in the steganographic domain: specifically, $\mathcal M$ is divided into a steganographic feature lifting network $\mathcal M_L$ and a FFD network $\mathcal M_D$. Meanwhile, during training, the steganographic domain alignment network extracts facial features ${\boldsymbol f}_{secret}=\mathcal M'_L(\boldsymbol x_{secret})$ to guide $\mathcal M$ toward latent facial information.
In this pipeline, only $\boldsymbol x_{stego}$ can be intercepted by attackers; therefore, $\boldsymbol x_{stego}$ must remain visually indistinguishable from $\boldsymbol x_{cover}$ to enable covert FFD without arousing suspicion.

\subsection{FFD in the Steganographic Domain}




We observe that existing FFD methods struggle to perform precise analysis within the steganographic domain. They often produce numerous false alarms (see Table~\ref{table:mainresult}, conventional detector), primarily because conspicuous information in the \emph{cover} image interferes with detection. 
We further notice that most semantic information of the cover is concentrated in the low-frequency band; meanwhile, in image hiding methods, information embedded in higher-frequency components is less likely to be detected than that in lower-frequency components.
Therefore, we propose a Low-Frequency-Aware Decomposition (LFAD) to extract the cover semantics and a Spatial-Frequency Differential Attention (SFDA) to suppress its interference from the stego image $\boldsymbol{x}_{stego}$, thereby enhancing the extraction of $\boldsymbol{x}_{secret}$.

\begin{figure*}[t]
    \centering
    \includegraphics[width=\textwidth]{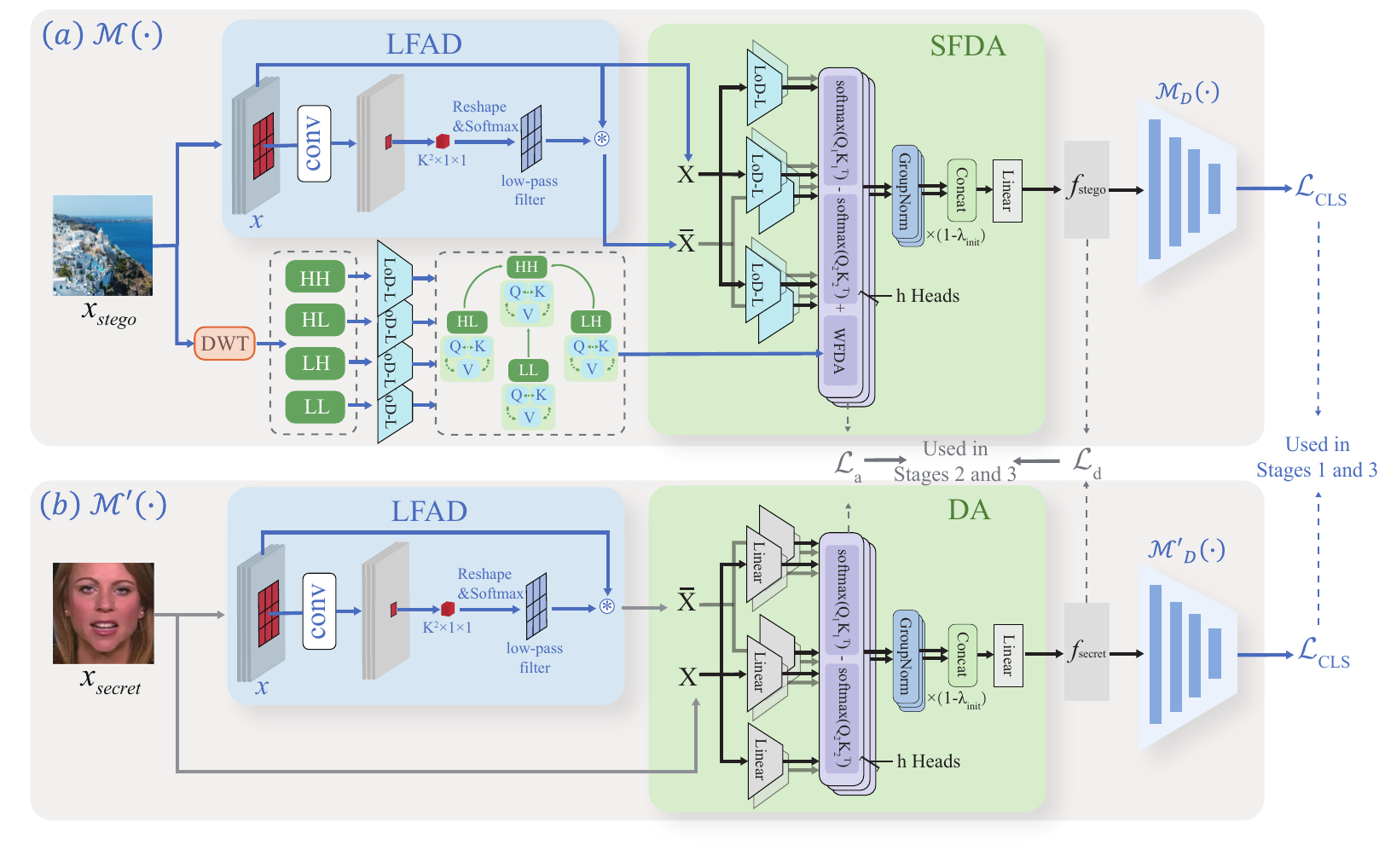}
    \caption{{Illustration of (a) the server-side network $\mathcal M(\cdot)$ of the proposed LFAD and SFDA. The feature lifting network contains LFAD and SFDA, which extracts high-frequency steganographic features containing facial information. (b) SDA branch $\mathcal M'(\cdot)$ helps to lift secret facial features by aligning stego-domain features with raw facial features.}}
    \label{fig:Detector}
\end{figure*}

In our StegaFFD network, the steganographic feature lifting network $\mathcal{M}_L$ consists of LFAD and SFDA. As illustrated in Fig.~\ref{fig:Detector}, the stego image $\boldsymbol{x}_{stego}$ is first decomposed into low-frequency features $\boldsymbol{\bar{x}}$ by LFAD. Subsequently, by integrating $\boldsymbol{x}_{stego}$ with $\boldsymbol{\bar{x}}$, SFDA extracts steganographic-domain features $\boldsymbol{f}_{stego}$ that capture the concealed information. 
Finally, $\boldsymbol{f}_{stego}$ is processed by the FFD network $\mathcal{M}_D$ to produce the final forgery detection result $\boldsymbol{\hat{y}}$. 

These operations effectively capture secret information, eliminate the influence of the cover image, and improve face forgery detection accuracy within the steganographic domain.
In the following sections, we provide detailed descriptions of the proposed LFAD and SFDA.

\paragraph{Low-Frequency-Aware Decomposition}


To estimate the cover-level feature, we construct spatially variant low-pass filters to effectively smooth high-frequency components~\cite{luo2022frequency}. We propose a Low-Frequency-Aware Decomposition (LFAD) network. Leveraging the advantages of high-level features~\cite{lu2022fade} to achieve high-quality adaptive low-pass filters, the LFAD network takes the hierarchical features $x$ extracted by the Xception feature network from the stego image $\boldsymbol{x}_{stego}$ as input and predicts spatial-variant low-pass filters. It comprises a $3\times 3$ convolutional layer followed by a kernel-wise softmax, represented as:
\begin{equation}
    {\bar V^l} = {\rm{Conv}}_{3 \times 3}(\boldsymbol{x}),
    ~\bar W_{i,j}^{l,p,q} = {\rm{Softmax}}(\bar V_{i,j}^l)
    = \displaystyle \frac{\exp\!\big(\bar V_{i,j}^{l,p,q}\big)}{\sum\limits_{(p',q')\in \Omega}\exp\!\big(\bar V_{i,j}^{l,p',q'}\big)}\!,
    \label{eq:2}
\end{equation}
where ${\bar V^l} \in \mathbb{R}^{\bar K^2 \times H \times W}$ stores, at each spatial location $(i,j)$, $\bar K^2$ logits corresponding to a $\bar K \times \bar K$ kernel; $\Omega$ denotes the index set of offsets within a $\bar K \times \bar K$ window. The kernel-wise softmax constrains the filter weights to be non-negative and sum to one, yielding smooth low-pass filters $\bar W \in \mathbb{R}^{\bar K^2 \times H \times W}$~\cite{zou2023delving}. Consequently, the low-pass filtered features $\bar x \in \mathbb{R}^{C \times H \times W}$ are obtained as:
\begin{equation}
    \boldsymbol{\bar x}_{i,j}^{\,l} = \sum_{(p,q)\in \Omega} \bar W_{i,j}^{l,p,q} \cdot \boldsymbol{x}_{i+p,\,j+q}^{\,l},
    \label{eq:3}
\end{equation}
where the operation is applied channel-wise for all $C$ channels (with symmetric padding at boundaries).

\paragraph{Spatial-Frequency Differential Attention}
%

To adaptively remove the cover information from the steganographic image $\boldsymbol{x}_{stego}$, we propose Spatial-Frequency Differential Attention (SFDA) as a foundation architecture. Differential attention takes the difference between different softmax attention functions to eliminate cover information and attention noise. Another advantage of this method is its consistency with differential amplifiers~\cite{laplante2018comprehensive} proposed in electrical engineering, where the difference between two signals is used as output, so that we can null out the common-mode noise of the input. 

We take a decoder-only model as an example to describe the architecture. Given an stego image $\boldsymbol x_{stego}$ and low-pass filtered features $\boldsymbol{\bar x}$, we pack the input embeddings into $\boldsymbol{X} \in \mathbb R^{ (H \times W) \times C}$ and $\boldsymbol{\bar X} \in \mathbb R^{(H \times W) \times C}$. The input is further contextualized to obtain the output $\boldsymbol f_{stego}$, i.e., $\boldsymbol f_{stego} = {\rm{Decoder}}(\boldsymbol{X}, \boldsymbol{\bar X})$. The decoder consists of two modules: a differential attention module followed by a feed-forward network module. Compared to vanilla Transformer~\cite{vaswani2017attention,morformer,hao2025position}, the main difference is the replacement of conventional softmax attention with differential attention, while the macro layout is kept the same. We also adopt pre-RMSNorm~\cite{zhang2019root} as improvements following LLaMA~\cite{touvron2023llama}.

The differential attention mechanism maps query, key, and value vectors to outputs. We use query and key vectors to compute attention scores, and then compute a weighted sum of value vectors. The critical design is that we use a pair of softmax functions to extract secret information and cancel the noise of attention scores. 

Given input features \( \boldsymbol{X} \) and \( \boldsymbol{\bar{x}} \) (flattened to the same shape as \(\boldsymbol{X}\)), we project them into query, key, and value representations using linear transformations:
\begin{equation}
    \begin{aligned}
        Q_1= \boldsymbol{X} W^Q,~Q_2 = \boldsymbol{\bar{x}} W^{LQ},~
        K_1 = \boldsymbol{X} W^K,~ K_2 = \boldsymbol{\bar{x}} W^{LK},~
        V = \boldsymbol{X} W^V,
    \end{aligned}
    \label{eq:qkv_inputs}
\end{equation}
where \( Q_1, Q_2, K_1, K_2, V \in \mathbb{R}^{(H \times W) \times d} \), and all \( W \) are learnable projection matrices. We apply pre-normalization (RMSNorm) to token embeddings before projections to stabilize training.

To capture more comprehensive frequency information, we further decompose \( X \) using the discrete wavelet transform (DWT) into four sub-bands:
\begin{equation}
    \boldsymbol{x}^{LL}, \boldsymbol{x}^{LH}, \boldsymbol{x}^{HL}, \boldsymbol{x}^{HH} = \mathrm{DWT}(\boldsymbol{X}), \quad \boldsymbol{x}^{(\cdot)} \in \mathbb{R}^{\left(\frac{H}{2} \times \frac{W}{2}\right) \times C}.
    \label{eq:DWT}
\end{equation}
Each sub-band is projected separately into queries and keys:
\begin{equation}
    \begin{aligned}
        Q_{hh} &= \boldsymbol{x}^{HH} W^{hhQ}, ~ Q_{hl} = \boldsymbol{x}^{HL} W^{hlQ}, 
        ~ Q_{lh} = \boldsymbol{x}^{LH} W^{lhQ}, ~ Q_{ll} = \boldsymbol{x}^{LL} W^{llQ}, \\
        K_{hh} &= \boldsymbol{x}^{HH} W^{hhK}, ~ K_{hl} = \boldsymbol{x}^{HL} W^{hlK}, 
        ~ K_{lh} = \boldsymbol{x}^{LH} W^{lhK}, ~ K_{ll} = \boldsymbol{x}^{LL} W^{llK}.
    \end{aligned}
    \label{eq:qkv_wavelet}
\end{equation}
The wavelet frequency-differential attention is computed as:
\begin{equation}
    \begin{aligned}
        \mathrm{WFDA}(\boldsymbol{X}) =\ 
        &\mathrm{softmax}\!\left(\frac{Q_{hh} K_{hh}^\top}{\sqrt{d}}\right)
        + \mathrm{softmax}\!\left(\frac{Q_{hl} K_{hl}^\top}{\sqrt{d}}\right) \\
        &- \mathrm{softmax}\!\left(\frac{Q_{lh} K_{lh}^\top}{\sqrt{d}}\right)
        - \mathrm{softmax}\!\left(\frac{Q_{ll} K_{ll}^\top}{\sqrt{d}}\right).
    \end{aligned}
    \label{eq:DiffAttn_wavelet}
\end{equation}
The final differential attention operator is defined as:
\begin{equation}
    \begin{aligned}
        \mathrm{DiffAttn}(\boldsymbol{X}, \boldsymbol{\boldsymbol{\bar{x}}}) =\ 
        \mathrm{softmax}\!\left(\frac{Q_1 K_1^\top}{\sqrt{d}}\right)
        - \lambda\, \mathrm{softmax}\!\left(\frac{Q_2 K_2^\top}{\sqrt{d}}\right) + \mathrm{WFDA}(\boldsymbol{X}),
    \end{aligned}
    \label{eq:DiffAttn_final}
\end{equation}
Here, \( W^Q, W^{LQ}, W^K, W^{LK}, W^V \in \mathbb{R}^{C \times d} \) are learnable projection matrices, and \( \lambda \) is a learnable scalar. To facilitate stable optimization, we re-parameterize \( \lambda \) as:
\begin{equation}
    \lambda = \exp(\lambda_{q_1} \cdot \lambda_{k_1}) - \exp(\lambda_{q_2} \cdot \lambda_{k_2}) + \lambda_{\mathrm{init}},
    \label{eq:lambda}
\end{equation}
where \( \lambda_{q_1}, \lambda_{k_1}, \lambda_{q_2}, \lambda_{k_2} \in \mathbb{R}^d \) are learnable vectors, and \( \lambda_{\mathrm{init}} \in (0, 1) \) is a constant initialization value. In practice we also clamp $\lambda$ into $[0,1]$ to prevent numerical explosion; setting \( \lambda_{\mathrm{init}} = 0.8 \) works well empirically.
%

%
To mitigate the loss of effective resolution caused by averaging attention weights across spatial positions, we further adopt the multi-head attention mechanism~\cite{vaswani2017attention} in our Differential Attention module. Let \( h \) denote the number of attention heads. For each head \( i \in [1, h] \), we use separate projection matrices \( W^Q_i, W^{LQ}_i, W^K_i, W^{LK}_i, W^V_i \), while sharing the same scalar \( \lambda \) across all heads within a layer. The output of each head is normalized and aggregated as follows:
\begin{equation}
    \begin{aligned}
    &\text{head}_i = \mathrm{DiffAttn}(\boldsymbol{X}, \boldsymbol{\bar{x}}; W^Q_i, W^{LQ}_i, W^K_i, W^{LK}_i, \lambda)\, V_i, \\
    &\overline{\text{head}_i} = (1 - \lambda_{\text{init}}) \cdot \mathrm{LN}(\text{head}_i), \\
    &\mathrm{MultiHead}(\boldsymbol{X}, \boldsymbol{\bar{x}}) = \mathrm{Concat}(\overline{\text{head}_1}, \ldots, \overline{\text{head}_h})\, W^O,
    \end{aligned}
    \label{eq:multihead}
\end{equation}
where \( V_i = \boldsymbol{X} W^V_i \in \mathbb{R}^{(H \times W) \times d} \), \( \mathrm{LN}(\cdot) \) applies RMSNorm~\cite{zhang2019root} to each head independently, and \( \mathrm{Concat}(\cdot) \) denotes concatenation along the channel dimension. We apply a fixed scaling factor \( (1 - \lambda_{\text{init}}) \) to RMSNorm outputs to maintain gradient alignment with standard Transformer architectures. We set the number of heads as \( h = \frac{H \times W}{2d} \), where \( d \) is the per-head feature dimension consistent with standard Transformer settings. The output projection matrix is \( W^O \in \mathbb{R}^{(h\cdot d) \times C} \).


As illustrated in Fig.~\ref{fig:Detector}, we use $\rm {GroupNorm}(\cdot)$~\cite{wu2018group} emphasize that $\rm {LN}(\cdot)$ is applied to each head independently. As differential attention tends to have a sparser pattern, statistical information is more diverse between heads. The $\rm {LN}(\cdot)$ operator normalizes each head before concatenation to improve gradient statistics~\cite{wang2023magneto,qin2022devil, ma2025decoupled}.

Finally, we describe the architecture of the Multi-Head Differential Attention network as: 
\begin{equation}
    \begin{array}{l}
    \boldsymbol f_{stego} = \lambda_{d} \cdot \rm {MultiHead}(LN(\boldsymbol{X},\boldsymbol{\bar X}))+\boldsymbol{X},
    \end{array}
    \label{eq:SFDA}
\end{equation}
where $\rm {LN}(\cdot)$ is RMSNorm~\cite{zhang2019root}, $\lambda_{d}$ is hyperparameter. 
Here LN$(\boldsymbol{X},\boldsymbol{\bar X})$ denotes applying normalization after concatenating $\boldsymbol{X}$ and $\boldsymbol{\bar X}$ along channels.

\subsection{Steganographic Domain Alignment}


%

\noindent To guide the $\mathcal{M}_L(\boldsymbol{x}_{stego})$ network to retain more information about $\boldsymbol{x}_{secret}$, we design an alignment method.
Steganographic Domain Alignment (SDA) network extracts the steganographic-domain features of the secret image, denoted as $\boldsymbol{f}_{secret}$, and guides $\mathcal{M}_L(\boldsymbol{x}_{stego})$ to align with $\boldsymbol{f}_{secret}$ when extracting the steganographic-domain features of the stego image. Notably, the SDA network is only involved during the training phase.
\begin{equation}
    \boldsymbol f_{secret}=\mathcal{M'}_L(\boldsymbol x_{secret}).
    \label{eq:fsecret}
\end{equation}
$\mathcal{M'}_L$ network is also a decoder-only model. Its attention matrix is denoted as 
\begin{equation}
\begin{aligned}
    &\boldsymbol{\bar x}_{secret} = {\rm{LFAD}}(\boldsymbol{x}_{secret}), \\
    &\rm {DiffAttn}(\boldsymbol{x}_{secret},\boldsymbol{\bar x}_{secret}) 
    = {\rm {softmax}}\Big(\frac{Q_1K_1^{\top}}{\sqrt{d}}\Big)\;
      - \lambda\cdot {\rm {softmax}}\Big(\frac{Q_2K_2^{\top}}{\sqrt{d}}\Big),
\end{aligned}
    \label{eq:sdaattention}
\end{equation}
where $X=\boldsymbol{x}_{secret}$, $\boldsymbol{\bar{x}}=\boldsymbol{\bar{x}}_{secret}$, and $Q_1,Q_2,K_1,K_2$ are computed as in Eq.~(\ref{eq:qkv_inputs}) (row-wise softmax). We use the same $\lambda$ re-parameterization as Eq.~(\ref{eq:lambda}). We note this attention network as Differential Attention (DA).

\paragraph{Steganographic Domain Distance Metric}
To align features in the steganographic domain, we must choose an appropriate distance between the stego and secret features. In domain adaptation, two families of metrics are commonly used: (i) moment-estimation criteria, such as MMD~\cite{gretton2012kernel}, CORAL~\cite{sun2016deep}, and HoMM~\cite{chen2020homm}; and (ii) information-theoretic divergences comparing distributions, including KL~\cite{perez2008kullback}, JS~\cite{jose2021information}, and Wasserstein~\cite{ruschendorf1985wasserstein}. 

Considering both training stability and the subtle distribution shift introduced by information embedding, we adopt CORAL as the core of SDA, as it is effective at capturing slight discrepancies~\cite{xue2023adaptive}. At the same time, because MMD primarily reflects the mean shift between domains, we use it as a dynamic coefficient to modulate the overall alignment strength. Concretely, our SDA distance is
\begin{equation}
\label{eq:sda_metric}
d_{SDA}(\boldsymbol f_{stego},\,\boldsymbol f_{secret})
= d_C(\boldsymbol f_{stego},\,\boldsymbol f_{secret})\cdot
\exp\!\big(\gamma\, d_M(\boldsymbol f_{stego},\,\boldsymbol f_{secret})\big),
\end{equation}
where \(d_C\) and \(d_M\) denote the standard CORAL and MMD distances, respectively; \(\gamma\) is a fixed hyperparameter (we set \(\gamma=10\)). Here, \(\boldsymbol f_{secret}\) is the feature lifted from the feature lifting network.

The loss function for aligning the steganographic domain is designed as
\begin{equation}
\label{eq:Ld}
\mathcal{L}_{d} = d_{SDA}(\boldsymbol f_{stego},\,\boldsymbol f_{secret}).
\end{equation}

\paragraph{{Low-rank} Decomposition fine-tuning}

%
%
Our experiments reveal that directly applying the $\mathcal{L}_{SDA}$ loss can sometimes degrade the recognition accuracy of the FFD network, suggesting that the alignment branch interferes with already learned representations. To minimize disruption to the well-learned semantic knowledge, inspired by~\cite{DBLP:conf/iclr/Ye0XSZ0W25}, we incorporate {Low-rank} Decomposition (LoD) to explicitly separate an orthogonal \emph{semantic} subspace from a \emph{forgery-specific} subspace, thereby preventing distortion of acquired semantics during steganographic domain alignment.

We obtain a rank-$r$ approximation of the pre-trained weight matrix with LoD in stage~1 (\ref{Training Process}); that is, we retain only the top $r$ singular values and the associated singular vectors:
\begin{equation}
    W_r = U_r \Sigma_r V_r^{\top},
    \label{eq:wr}
\end{equation}
where $U_r \in \mathbb{R}^{n \times r}$, $\Sigma_r \in \mathbb{R}^{r \times r}$, and $V_r \in \mathbb{R}^{n \times r}$. We keep $W_r$ frozen throughout subsequent training stages to preserve knowledge learned from stage 1 training. 

The residual component is defined as the difference between the full weights and the LoD approximation: 
\begin{equation}
    \Delta W \;=\; W - W_r \;=\; U_{n-r}\,\Sigma_{n-r}\,V_{n-r}^{\top},
    \label{eq:deltaW}
\end{equation}
where $U_{n-r} \in \mathbb{R}^{n \times (n-r)}$, $\Sigma_{n-r} \in \mathbb{R}^{(n-r) \times (n-r)}$, and $V_{n-r} \in \mathbb{R}^{n \times (n-r)}$. Note that $\Delta W$ is the learnable part, corresponding to the remaining singular components and reflecting slight, targeted adjustments to the original weight matrix.

During training, we optimize only $\Delta W$ while keeping $U_r$, $\Sigma_r$, and $V_r$ fixed. This implementation preserves stage~1 semantics via $W_r$ and adapts the model to align the steganographic domain through the residual $\Delta W$. We evaluate $n-r \in \{1,4,16,64\}$ and observe that $n-r=16$ yields the best performance.

\begin{algorithm}[t]
\small
\caption{Training Process of StegaFFD}
\label{alg:train}
\begin{algorithmic}[1]
\Require DIH network $\mathcal H$, SDA networks $(\mathcal M'_L,\mathcal M'_D)$, StegaFFD feature lifting network $\mathcal M_L$, FFD network $\mathcal M_D$, training set $\mathcal T$, stage epochs $(e_1,e_2,e_3)$, learning rates $(\ell_r^1,\ell_r^2,\ell_r^3)$
\State $E \gets e_1+e_2+e_3$
\For{$j=1$ to $E$}
  \For{each $(\boldsymbol x_{cover}, \boldsymbol x_{secret}, \boldsymbol y)\in\mathcal T$}
    \State $\boldsymbol{x}_{stego} \leftarrow \mathcal H(\boldsymbol x_{cover}, \boldsymbol x_{secret})$
    \State $\boldsymbol{\bar x} \leftarrow \mathrm{LFAD}(\boldsymbol{x}_{stego})$
    \State $\bar X, X \leftarrow \boldsymbol{\bar x}, \boldsymbol{x}_{stego}$
    \State $\boldsymbol{f}_{stego} \leftarrow \mathrm{SFDA}(\bar X, X)$
    \State $\boldsymbol{f}_{secret} \leftarrow \mathcal M'_L(\boldsymbol{x}_{secret})$
    \If{$j \le e_1$} \Comment{\textcolor{teal}{Stage 1: train $\mathcal M,\mathcal M'$ with $\mathcal L_{CLS}$; freeze $\mathcal H$}}
        \State $\hat{\boldsymbol y} \leftarrow \mathcal M_D(\boldsymbol{f}_{stego})$, \quad
               $\hat{\boldsymbol y}' \leftarrow \mathcal M'_D(\boldsymbol{f}_{stego})$
        \State $\mathcal L_{CLS} \leftarrow \text{Eq.\,\eqref{eq:cls}}$
        \State $\theta_{\mathcal M,\mathcal M'}\leftarrow \theta_{\mathcal M,\mathcal M'}-\ell_{r}^1\cdot\nabla_{\theta_{\mathcal M,\mathcal M'}}\mathcal L_{CLS}$
    \ElsIf{$e_1 < j \le e_1+e_2$} \Comment{\textcolor{teal}{Stage 2: align $\mathcal M_L$ with $\boldsymbol f_{secret}$; freeze $\mathcal H$}}
        \State $\mathcal L_{d} \leftarrow \text{Eq.\,\eqref{eq:sda_metric}}$, \quad
               $\mathcal L_{a} \leftarrow \text{Eq.\,\eqref{eq:la}}$
        \State $\mathcal L_{SDA} \leftarrow \mathcal L_{d}+\,\mathcal L_{a}$
        \State $\theta_{\mathcal M_L}\leftarrow \theta_{\mathcal M_L}-\ell_{r}^2\cdot\nabla_{\theta_{\mathcal M_L}}\mathcal L_{SDA}$
    \Else \Comment{\textcolor{teal}{Stage 3: joint fine-tuning of $\mathcal M$ and $\mathcal H$ with $\mathcal L_{total}$}}
        \State $\hat{\boldsymbol y} \leftarrow \mathcal M_D(\boldsymbol{f}_{stego})$, \quad
               $\hat{\boldsymbol y}' \leftarrow \mathcal M'_D(\boldsymbol{f}_{secret})$
        \State $\mathcal L_{CLS} \leftarrow \text{Eq.\,\eqref{eq:cls}}$, \quad
               $\mu \leftarrow \text{Eq.\,\eqref{eq:mu}}$
        \State $\mathcal L_{total} \leftarrow \mathcal L_{CLS}+\mu\,\mathcal L_{SDA}$ \Comment{\textcolor{teal}{See Eq.\,\eqref{eq:total}}}
        \State $\theta_{\mathcal M}\leftarrow \theta_{\mathcal M}-\ell_{r}^3\cdot\nabla_{\theta_{\mathcal M}}\mathcal L_{total}$
        \State $\theta_{\mathcal H}\leftarrow \theta_{\mathcal H}-\ell_{r}^3\cdot\nabla_{\theta_{\mathcal H}}\mathcal L_{total}$
    \EndIf
  \EndFor
\EndFor
\end{algorithmic}
\end{algorithm}

\paragraph{Attention Alignment}

To align the domain-invariant features extracted by the StegaFFD feature lifting network, we update the network by minimizing the distribution distance between the stego and secret features. In addition, to further couple the information captured by the feature lifting network and SDA branches, we introduce an attention-alignment term. For each stega–secret pair, the loss is
\begin{equation}
\begin{aligned}
    \mathcal{L}_{a} 
    &= \big\|\boldsymbol A_{stego}-\boldsymbol A_{secret}\big\|_F^{2}, 
    \label{eq:la}
\end{aligned}
\end{equation}
where $\boldsymbol A_{stego} = {\rm{DiffAttn}}(X,\,\bar X)$, and $\boldsymbol A_{secret} = {\rm{DiffAttn}}(\boldsymbol{x}_{secret},\,\boldsymbol{\bar x}_{secret})$.

Finally, we present the whole loss function for the alignment of the steganographic domain
\begin{equation}
\begin{aligned}
    \mathcal{L}_{SDA} 
    &= \mathcal{L}_{d} + \mathcal{L}_{a} .
    \label{eq:sda}
\end{aligned}
\end{equation}

\subsection{Training Process} \label{Training Process}
\noindent We use a multi-stage training strategy for StegaFFD, as shown in Algorithm~\ref{alg:train}.
\textbf{Stage~1:} We freeze the pretrained DIH network and train the facial image analysis network (together with the SDA branch) using the classification loss $\mathcal{L}_{CLS}$. 
\textbf{Stage~2:} We continue freezing the DIH network and \emph{align} lifted features with $\boldsymbol{f}_{secret}$ by minimizing $\mathcal{L}_{SDA}$. 
\textbf{Stage~3:} We fine-tune the whole StegaFFD (i.e., the DIH and facial image analysis networks) under the total loss $\mathcal{L}_{total}$.

\begin{equation}
\begin{aligned}
    \mathcal{L}_{CLS} 
    &= -\frac{1}{N_b}\sum_{i=1}^{N_b}\!\Big[
        \boldsymbol{y}_i\,\log(\hat{\boldsymbol{y}}_i)
        +(1-\boldsymbol{y}_i)\,\log\!\big(1-\hat{\boldsymbol{y}}_i\big) \\
    &\hspace{4.2em}
        +\boldsymbol{y}_i\,\log(\hat{\boldsymbol{y}}'_{i})
        +(1-\boldsymbol{y}_i)\,\log\!\big(1-\hat{\boldsymbol{y}}'_{i}\big)
    \Big].
\end{aligned}
\label{eq:cls}
\end{equation}

\noindent The total loss in model training is then defined as:
\begin{equation}
    \mathcal{L}_{total} = \mathcal{L}_{CLS}+\mu\cdot \mathcal{L}_{SDA},
    \label{eq:total}
\end{equation}
where $\mu$ is a hyperparameter that balances the two losses. During training, $\mu$ is gradually increased with the number of iterations so that the cross-entropy term dominates early training and helps learn discriminative stego features:
\begin{equation}
    \mu = \frac{2}{1+\exp(-\gamma_s \cdot \frac{s_t}{s_T})}-1,
    \label{eq:mu}
\end{equation}

where $s_t$ denotes the current training step, $s_T$ represents the total number of training steps and $\gamma_s$ is a hyperparameter, set to $\gamma_s=10$ in our experiments. The range of $\mu$ gradually increases from 0 to 1 over the course of training.

\section{Experiment}
\subsection{Experimental Setup}

\noindent \textbf{Datasets.}
For comprehensive comparisons, we use a collection of {eight} widely recognized and extensively used datasets in the realm of face forgery detection: {FaceForensics++~\cite{rossler2019faceforensics++}, CelebDF-v1~\cite{li2020celeb}, CelebDF-v2~\cite{li2020celeb}, DeepFakeDetection~\cite{rana2022deepfake}, DeepFake Detection Challenge Preview~\cite{dolhansky2019deepfake}, DeepFake Detection Challenge~\cite{dolhansky2020deepfake}, UADFV~\cite{li2018ictu} and FaceShifter~\cite{li2020advancing}. 
FaceForensics++ consists of 1,000 real and 4,000 fake videos generated by {four} distinct deepfake synthesis methods.
CelebDF-v1 (CDFv1) contains 408 real and 795 fake videos primarily created using one high-quality deepfake method.
CelebDF-v2 (CDFv2) is an upgraded version of v1, it features 590 real and 5,639 fake videos forged with an improved version of a single method.
DeepFakeDetection (DFD) includes 363 real and 3,000 fake videos generated from five different synthesis techniques.
DeepFake Detection Challenge Preview (DFDCP) contains 1,131 real and 4,119 fake videos from two methods, it also incorporates three types of perturbations.
DeepFake Detection Challenge (DFDC) is a massive dataset with 23,654 real and 104,500 fake videos produced by eight methods and includes 19 perturbations.
UADFV is a smaller-scale dataset, it consists of 49 real and 49 fake videos created using one deepfake method.
FaceShifter (Fsh) contains 1,000 real and 1,000 fake videos forged using a single, high-fidelity face-swapping technique.}
FaceForensics++ is employed for model training, while the rest are frequently used as testing data.

\vspace{2mm}
\noindent\textbf{Competing Detectors.}
To the best of our knowledge, no method has been proposed to perform FFD in the steganographic domain. To ensure a comprehensive and fair comparison, we select a \textbf{face anonymization method} Falco~\cite{barattin2023attribute} and four widely used \textbf{DIH methods} (i.e., WengNet~\cite{weng2019high}, HiDDeN~\cite{zhu2018hidden}, BalujaNet~\cite{baluja2019hiding} and HiNet~\cite{jing2021hinet}) to perform privacy-preserving. Then we deploy four famous or SOTA face forgery detection networks, i.e., Xception~\cite{rossler2019faceforensics++}, Meso4~\cite{afchar2018mesonet}, MesoIncep~\cite{afchar2018mesonet}, and F3Net~\cite{qian2020thinking}. We pair each privacy-preserving method with each FFD network, forming {twenty} competing methods.

\vspace{2mm}
\noindent\textbf{Evaluation Metrics.}
Regarding evaluation, we compute the average value of the top-3 metrics, such as the average top-3 area under the curve (AUC). Additionally, we report the top-1 results. Other widely used metrics, including accuracy (ACC), average precision (AP), and equal error rate (EER), are also computed and presented in the following sections.

\vspace{2mm}
\noindent\textbf{Implementation Details.}
In the data processing, face detection, face cropping, and alignment are performed using DLIB~\cite{sagonas2016300}. The aligned faces are resized to 256 × 256 for both the training and testing. We also apply widely used data augmentations, i.e., image compression, horizontal flip, rotation, Gaussian blur, and random brightness contrast. In the training module, we use the Adam optimizer with learning rates $\ell_r^1$, $\ell_r^2$, $\ell_r^3$ of $2 \times 10^{-4}$,$1 \times 10^{-4}$,$1 \times 10^{-5}$, respectively. The batch size is fixed at 32 for all experiments. We set $\lambda_{init}$, $\lambda_{d}$  of SFDA shown in Eq. (\ref{eq:lambda},\ref{eq:SFDA}) to 0.8, 2. We primarily leverage pre-trained backbones from ImageNet if feasible. Otherwise, we resort to initializing the remaining weights using a normal distribution. During training, all cover images $\boldsymbol x_{cover}$ are randomly selected.

\subsection{Comparison Results}

\noindent \textbf{Classification Results.}
As shown in Table~\ref{table:mainresult}, StegaFFD achieves the highest detection performance across seven datasets compared to other covert FFD methods, respectively. On average, its segmentation performance surpasses that of the second-place Xception + HiNet by 5.16\% in AUC. When compared to the vanilla framework (without privacy-enhancing) utilizing Xception, StegaFFD's average performance only shows a slight decrease of 1.96\% in AUC. This indicates StegaFFD's effectiveness in maintaining high FFD performance while enhancing privacy protection. 
We note that F3Net also utilizes information from the frequency domain and extracts frequency artifacts for detection (similar to StegaFFD), but underperforms due to its lack of adaptive analysis across different frequencies of the stego image. 
Additionally, despite Meso4, MesoIncep, and F3Net having incremental designs over Xception, their performance does not improve across all datasets. 
This inconsistency can be attributed to the fact that their incremental designs are not effective in the steganographic domain.

\begin{table*}[t]
\renewcommand{\arraystretch}{1}
\setlength{\tabcolsep}{1pt} 
\caption{Cross-domain evaluations using the AUC {(\%)} metric. All detectors are trained on FF-c23 and evaluated on other data. ``Avg.'' denotes the average AUC for within-domain and cross-domain evaluation, and the overall results. ``Top-3'' represents the count of each method's ranks within the top-3 across all testing datasets. ``Freq.'' denotes the frequency type. {Privacy Preserving method include (i) unprotected (None), (ii) an anonymization method (Falco~\cite{barattin2023attribute}), (iii) existing steganography method (HiNet~\cite{jing2021hinet} to BalujaNet~\cite{baluja2019hiding}).} The best-performing method for each column is highlighted in \textcolor{red!80!black}{red} and the top three are shown in \textbf{bold}.    }
    
  \begin{center}

    \label{table:mainresult} 
\resizebox{1.\textwidth}{!}{
    \begin{tabular}{l|c|c|c c c c c c c| c| c} 
      \toprule  
      
      \multirow{2}*{\shortstack{ \textbf{Privacy}\\ \textbf{Preserving}}}& \multirow{2}*{\textbf{Type}} & \multirow{2}*{\shortstack{ \textbf{FFD}\\ \textbf{Detector}}}& \multicolumn{9}{c} {\textbf{Cross Domain Evaluation}} \\
      \cmidrule(lr){4-12}
      &  & &\textbf{CDFv1} & \textbf{CDFv2} & \textbf{DFD} & \textbf{DFDC}& \textbf{DFDCP}& \textbf{Fsh}& \textbf{UADFV}& \textbf{Avg.}& \textbf{Top-3}\\
      \midrule
      None & Plain & Xception & 74.85  & 75.97 &82.11 &69.19 &68.11 &59.48 &88.01 &73.96 &--- 
\\
      None & Plain & Meso4 & 63.69  & 55.65 &52.97 &58.89 &49.03 &52.90 &59.05 &56.03 &---
\\
      None & Plain & MesoIncep & 70.84  & 70.02 &59.80 &60.16 &74.57 &69.21 &89.78 &70.63  &---
\\
      None & Freq. & F3Net & 62.91  & 64.89 &83.88 &66.61 &70.84 &72.66 &77.76 &71.36  &---
\\

     \midrule
      Falco  & Plain & Xception & 69.12  & 65.41 &59.49 &58.54 &61.92 &54.98 &51.48 &60.13 &0 
\\
      Falco  & Plain & Meso4 & 45.35  & 55.93 &49.23 &50.82 &55.81 &54.06 &53.10 &52.04  &0
\\
      Falco  & Plain & MesoIncep & 58.34  & 62.56 &50.15 &50.77 &61.40 &54.43 &60.48 &56.88   &0
\\
      Falco  & Freq. & F3Net & 56.53  & 59.04 &56.59 &52.91 &62.32 &54.54 &50.01 &55.99  &0
      \\
      \midrule
      HiNet  & Plain & Xception & 74.94  & 66.94 &70.40 &63.85 &64.26 &47.01 &80.49 &\textbf{66.84}  & 0
\\
      HiNet  & Plain & Meso4 & 64.11  & 60.60 &50.61 &52.09 &55.29 &50.38 &53.01 &55.16  &0
\\
      HiNet  & Plain & MesoIncep & 57.78  & 62.56 &58.77 &57.86 &\textcolor{red!80!black}{\textbf{71.73}} &{\textbf{62.15}} &\textcolor{red!80!black}{\textbf{78.21}} &64.15   &3 
\\
      HiNet  & Freq. & F3Net & 72.45  & 71.08 &71.83 &\textbf{63.40} &60.59 &53.13 &68.04 &65.79   &1
\\
      HiDDeN  & Plain & Xception & 68.96  & 64.27 &72.07 &{\textbf{63.80}} &64.45 &49.64 &\textbf{74.36} &65.51   &2
\\
      HiDDeN  & Plain & Meso4 & 63.93  & 64.65 &53.32 &56.13 &57.90 &56.94 &76.31 &61.31   &0
\\
      HiDDeN  & Plain & MesoIncep & 65.64  & 67.99 &58.34 &60.47 &\textbf{69.25} &{\textbf{61.72}} &73.56 &65.28 &2
\\
      HiDDeN  & Freq. & F3Net & 68.80  &69.55 &69.02 &63.13 &62.42 &57.15 &68.24 &65.47  &0
\\
      WengNet  & Plain & Xception & 67.70  & 66.23 &70.5 &64.01 &65.74 &50.44 &75.81 &65.78   &0
\\
      WengNet  & Plain & Meso4 & 51.66  & 58.16 &51.89 &52.23 &59.16 &53.64 &60.80 &55.36   &0
\\
      WengNet  & Plain & MesoIncep & 60.07  & 57.47 &52.48 &50.74 &52.98 &50.23 &52.14 &53.73  &0
\\
      WengNet  & Freq. & F3Net & \textbf{76.40}  & {\textbf{72.17}} &70.38 &60.86 &62.52 &53.35 &69.78 &\textbf{66.49}  &2
\\
      BalujaNet  & Plain & Xception & 70.56 & 68.90 &75.55 &62.61 &63.15 &50.75 &\textbf{74.05} &65.22 &1
\\
      BalujaNet  & Plain & Meso4 & 62.48  & 62.40 &50.70 &52.68 &58.86 &50.85 &66.14 &57.73   &0
\\
      BalujaNet  & Plain & MesoIncep & 66.24  & 64.05 &61.52 &61.41 &66.96 &\textcolor{red!80!black}{\textbf{63.50}} &75.47 &65.59  & 1
\\
      BalujaNet  & Freq. & F3Net & 60.48  & 65.93 &76.36 &61.82 &66.88 &58.52 &65.32 &65.04 &0
\\
 \midrule
      HiDDeN  & Freq. & SFDA & 72.57 & 66.20 &\textbf{80.23} &55.20 &68.85 &53.04 &66.05 &66.02 & 1
\\
      HiDDeN  & Freq. & SFDA+SDA & \textbf{77.15} & \textbf{72.41} &\textcolor{red!80!black}{\textbf{81.47}} &52.09 &65.78 &50.03 &50.04 &64.14 &3
\\
      HiDDeN  & Freq. & StegaFFD & \textcolor{red!80!black}{\textbf{81.22}} &\textcolor{red!80!black}{\textbf{72.43 }}&\textbf{80.64} &\textcolor{red!80!black}{\textbf{66.40}} &\textbf{71.03} &58.68 &73.63 &\textcolor{red!80!black}{\textbf{72.00}} &5
\\
      \bottomrule 
    \end{tabular}
    }
  \end{center}

\end{table*}

\begin{figure*}[t]
    \centering
    \includegraphics[width=0.9\textwidth]{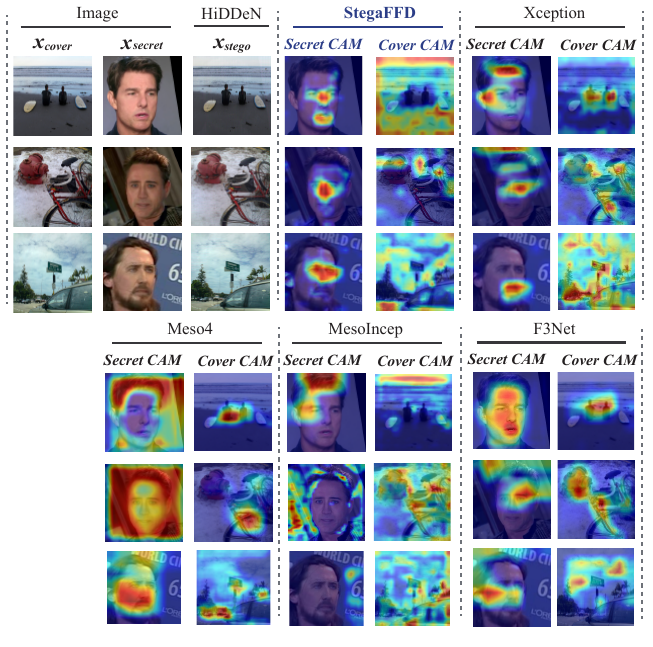}
    \caption{Visual attribution analysis of face forgery detection algorithms (i.e., StegaFFD, Xception~\cite{rossler2019faceforensics++}, Meso4~\cite{afchar2018mesonet}, MesoIncep~\cite{afchar2018mesonet}, and F3Net~\cite{qian2020thinking}) using Grad-CAM++, highlighting feature focus on cover and secret image regions (e.g., facial forgery areas).}
    \label{fig:Attribution_analysis}
\end{figure*}

\begin{figure}[t]
    \centering
    \includegraphics[width=.7\textwidth]{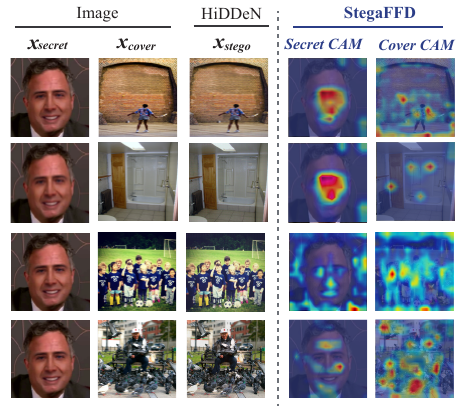}
    \caption{{Robust attribution analysis of face forgery detection across varied cover images for a single facial instance and failure cases. }This figure illustrates Grad-CAM++ attribution heatmaps highlighting feature focus on cover and secret image regions (e.g., facial forgery areas).}
    \label{fig:covers}
\end{figure}

\begin{figure}[t]
    \centering
    \includegraphics[width=.95\textwidth]{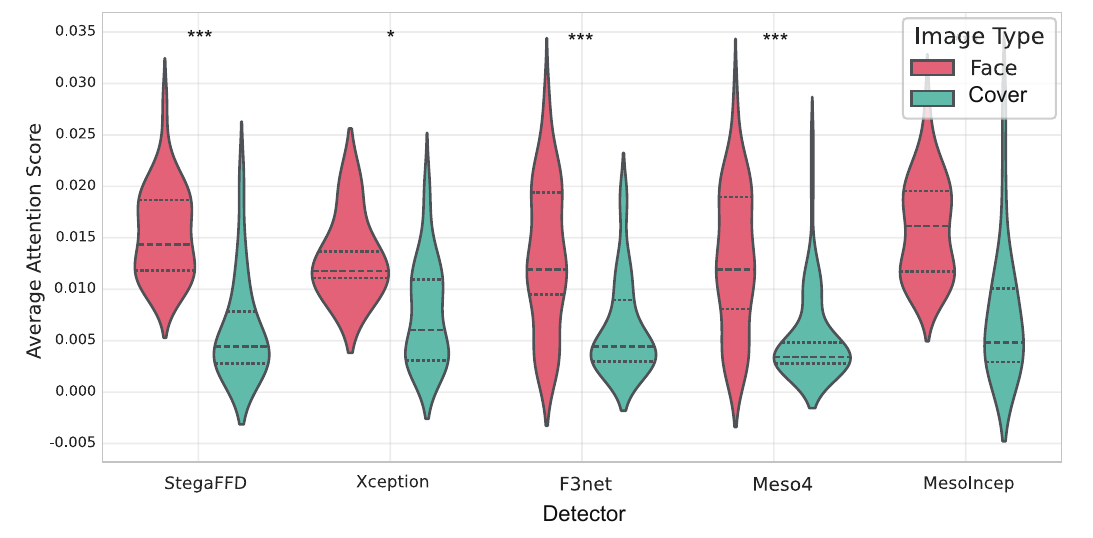}
    \caption{Quantitative attribution analysis using violin plots comparing attention to embedded face regions versus natural backgrounds across detectors. StegaFFD exhibits a markedly larger face–natural separation than the four baselines, indicating stronger, more consistent focus on face content.}
    \label{fig:violin}
\end{figure}

\vspace{2mm}
\noindent\textbf{Secret Sensibility.}
%
As shown in Fig.~\ref{fig:Attribution_analysis}, $\boldsymbol x_{stego}$ generated by DIH has the highest visual similarity to $\boldsymbol x_{cover}$, thereby enhancing facial privacy protection.
To demonstrate that the features extracted by StegaFFD are cover-agnostic, we conduct an attribution analysis on the feature space ($\boldsymbol f_{stego}$). Utilizing the GradCAM~\cite{DBLP:journals/ijcv/SelvarajuCDVPB20} algorithm, it is evident that StegaFFD's feature space ($\boldsymbol f_{stego}$) is primarily focused on facial forgery regions (e.g., eyes, nose, and mouth) while remaining insensitive to the main content of the cover image. In contrast, other algorithms either suffer from significant influence by the cover's semantic information (e.g., Meso4, F3Net) or fail to extract effective semantic features (e.g., MesoIncep).
The poor secret sensibility of competing methods is attributed to the absence of low-frequency guidance from $\boldsymbol x_{stego}$. To demonstrate the robustness of our algorithm across different natural cover images, we embed a single face image into four distinct natural cover images, as shown in Fig.~\ref{fig:covers}, with attribution results indicating that StegaFFD consistently focuses on the face region, exhibiting strong robustness.

We quantify the results of an attribution analysis to determine whether the detectors prioritize the embedded face content over the natural cover image in our steganographic setting (Fig.~\ref{fig:violin}). For StegaFFD and four baseline detectors, we generate per-image heatmaps from intermediate feature activations. We then calculate an attention score using a JET-colormap-aware, color-weighted metric, where higher red intensity and lower blue intensity reflect stronger attention. We analyze {twenty} pairs of face images and cover images. The data is categorized into ``face'' and ``cover'' sets for each algorithm and evaluated using the mean attention score as the primary metric, with quartile lines for additional robustness checks. Violin plots visualize the attention score distributions across algorithms. We apply two-sample t-tests to assess differences between face and cover attention groups.
Results show that StegaFFD exhibits a significantly larger separation between face and cover attention distributions compared to the baseline detectors. Its face attention scores are consistently higher, with stronger statistical significance, demonstrating that StegaFFD more effectively focuses on the embedded face content rather than the cover image.

{\noindent\textbf{Imperceptibility}
As shown in Fig.~\ref{fig:Attribution_analysis} and Fig.~\ref{fig:covers}, $\boldsymbol x_{stego}$ generated by DIH has the highest visual similarity to $\boldsymbol x_{cover}$, thereby enhancing facial privacy protection.
We also analyzed the similarity between $\boldsymbol x_{cover}$ and $\boldsymbol x_{stego}$, as well as between $\boldsymbol x_{secret}$ and $\boldsymbol x_{stego}$. We reported PSNR and SSIM to quantify their similarity. 
The results indicate that $\boldsymbol x_{cover}$ and $\boldsymbol x_{stego}$ are highly similar, demonstrating that StegaFFD has excellent imperceptibility (Table~\ref{table:Imperceptibility}).}

\begin{table}[t]
\caption{{Similarity metrics PSNR and SSIM for image pairs $\boldsymbol x_{cover}/\boldsymbol x_{stego}$ and $\boldsymbol x_{secret}/\boldsymbol x_{stego}$, demonstrating StegaFFD's imperceptibility.}}
\centering
    \begin{tabular}{l| c |c} 
      \toprule  
      \multirow{1}*{\enspace \textbf{Image Pair} }  &PSNR$\uparrow$ &SSIM$\uparrow$  \\
      \midrule
      $\boldsymbol x_{cover}/\boldsymbol x_{stego}$ &   32.46 &0.86  
\\
      $\boldsymbol x_{secret}/\boldsymbol x_{stego}$ & 10.38  & 0.41
\\
      \bottomrule 
    \end{tabular}
    \label{table:Imperceptibility}
\end{table}

\begin{figure}[t]
    \centering
    \includegraphics[width=.7\textwidth]{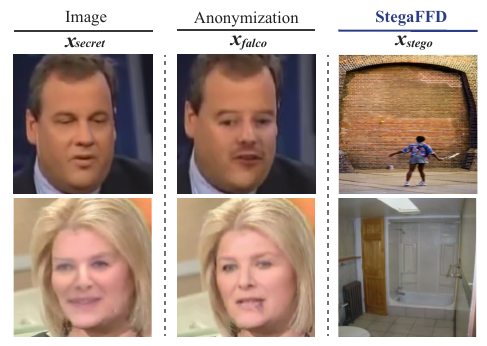}
    \caption{{Comparison of imperceptibility between StegaFFD and anonymization-based methods. StegaFFD hides facial information within natural images, achieving higher imperceptibility, while anonymization-based methods introduce face forgery features, leading to reduced FFD accuracy.}}
    \label{fig:anonymization}
\end{figure}

{To further compare the advantages of StegaFFD and anonymization-based methods in terms of imperceptibility, we conducted a separate comparison in Fig.~\ref{fig:anonymization}. Although anonymization-based methods modify facial identity information, they introduce face forgery features, resulting in a disadvantage in FFD accuracy. In contrast, StegaFFD hides facial information within natural images, making privacy information less perceptible.}

{\noindent\textbf{Failure Cases}
To better inform the community about the advantages and limitations of StegaFFD, and to avoid failures in practical applications, we analyzed the failure cases of StegaFFD. The last two rows of Fig.~\ref{fig:covers} showcase these failure cases. 
When the cover image contains densely packed objects, it may affect detection results. This is because our method focuses on extracting high-frequency information from steganographic images. If the cover image itself contains dense objects, it generates high-frequency information, which may interfere with the detector. Therefore, in practical use, selecting natural images with less dense objects as covers can help avoid potential issues.}

\begin{table}[t]
\caption{Ablation results for SFDA variants on CDFv1 and CDFv2 datasets. {All metrics in the table are presented as percentages.}}
\centering
    \begin{tabular}{l| cccc |cccc} 
      \toprule  
      \multirow{1}*{\enspace \textbf{Variation} }  &\multicolumn{4}{c} {\textbf{CDFv1}}&\multicolumn{4}{c} {\textbf{CDFv2}}  \\
      \cmidrule(lr){2-9}
      \textbf{ of SFDA } & AUC $\uparrow$ &ACC$\uparrow$&AP$\uparrow$&EER $\downarrow$& AUC $\uparrow$ &ACC$\uparrow$&AP$\uparrow$&EER $\downarrow$ \\
      \midrule
      LFAD &   53.04 &38.36  &54.05&47.38 & 58.18&34.22  &61.72   & 44.43
\\
      HFAD &  64.15 & 65.33 &62.84& 39.48& 60.44&  64.78 &65.64&40.10
\\
      w/o DWT &  73.20  &62.24 &81.57& 32.00&71.46 & 66.12  &73.51& 33.77
\\
      \cellcolor{gray!20}StegaFFD &   \cellcolor{gray!20}81.22  & \cellcolor{gray!20}71.25 & 
    \cellcolor{gray!20}86.93
    &\cellcolor{gray!20}24.85 &\cellcolor{gray!20}72.43  &\cellcolor{gray!20}70.68 
    &\cellcolor{gray!20}82.95
    &\cellcolor{gray!20}26.62
\\
      \bottomrule 
    \end{tabular}
    \label{table:SFDA}
\end{table}

\begin{table}[t]
\caption{Ablation study on SDA Loss formulations for StegaFFD on CDFv1 datasets. {All metrics in the table are presented as percentages.}}
\centering
    
    \begin{tabular}{c|ccc c} 
      \toprule  
       \textbf{SDA Loss }   & AUC $\uparrow$ &ACC$\uparrow$&AP$\uparrow$&EER $\downarrow$ \\
      \midrule
      None &  72.57  &62.21  & 79.04&31.92
\\
      FA(l2) loss &   73.20 & 63.99 &75.46 &34.33
\\
      AA(sda) loss  &    70.88   & 66.35 & 78.49& 36.24
\\
      FA(l2)+AA(l2) loss  & 75.31   & 63.23 & 81.96& 30.25
\\
      FA(sda)+AA(sda) loss  &  79.65   &65.63 & 84.35& 27.09
\\
      \cellcolor{gray!20}FA(l2)+AA(sda) loss  &  \cellcolor{gray!20}81.22  &\cellcolor{gray!20}71.25 &\cellcolor{gray!20}86.93 &\cellcolor{gray!20}24.85 
\\
      \bottomrule 
    \end{tabular}
    \label{table:SDA}
\end{table}

\subsection{Ablation Study} 
\label{Ablation Study}

\noindent \textbf{Effectiveness of SFDA.}
To validate the effectiveness of SFDA in extracting secret information, we deploy several variants of SFDA. 1) We directly apply low-frequency filters to extract the low-frequency signal $\boldsymbol{\bar{x}}$ as the feature $f_{stego}$, denoted as Direct LFAD. 2) We reformulated the filters $\bar{W}_{i,j}^{l,p,q}$ in Eq.~\ref{eq:2} as high-pass filters,
\begin{equation}
 \bar W^{p,q}_{i,j} = {\rm{E}}^{p,q} - {\rm{Softmax}}(\bar V_{i,j}^{l,p,q})= {\rm{E}}^{p,q} - \frac{\exp(\bar V^{\,l,p,q}_{i,j})}{\sum_{(p',q')\in\Omega} \exp(\bar V^{\,l,p',q'}_{i,j})}.
    \label{eq:HFAD}
\end{equation}
To ensure the final generated kernels $\bar W$ are high-pass, we follow [83] to invert the low-pass kernels by subtracting them from the identity kernel $\rm{E}$, whose weights are [[0, 0, 0], [0, 1, 0], [0, 0, 0]] when $\bar K = 3$. {$\Omega$ denotes the index set of offsets within a $\bar K \times \bar K$ window.} We note it as HFAD. 3) We remove the discrete wavelet transformation (DWT) from the differential attention mechanism, referred to as w/o DWT.

As shown in Table~\ref{table:SFDA}, Direct LFAD demonstrates that most of the semantic information in the cover, rather than the secret, is distributed within the low-frequency band. As a high-pass filter, HFAD is theoretically capable of extracting more secret information. The performance gap between HFAD and StegaFFD indicates that SFDA is more effective in extracting secret information. This may be attributed to HFAD's high-pass filtering discarding excessive information, whereas differential attention adaptively suppresses interfering information, achieving greater stability. Furthermore, SFDA w/o DWT suggests that the discrete wavelet transformation contributes to more comprehensive extraction of frequency information from the stego image $x_{stego}$.

\vspace{2mm}
\noindent\textbf{Effectiveness of SDA.}
Table~\ref{table:SDA} presents an ablation study on different loss formulations for SDA. The alignment information is categorized into Attention Alignment (AA) and Feature Alignment (FA). Based on the loss type, these are further divided into L2 loss and SDA loss, resulting in five comparative groups. The ``None'' group indicates the absence of any alignment loss. The combination of FA(L2)+AA(SDA) corresponds to the loss formulation used in StegaFFD. To provide a more comprehensive evaluation, we report the metrics of AUC, ACC, AP, and EER. The results demonstrate that SDA effectively assists the StegaFFD network in extracting more facial information, thereby improving FFD performance.

\vspace{2mm}
\noindent\textbf{Effectiveness of LoD.}
Table~\ref{table:LoD} validates the comparison between configurations with and without LoD. To provide a more comprehensive evaluation, we report the metrics of AUC, ACC, AP, and EER. Experiments were conducted on the CDFv1, CDFv2, and UADFV datasets to ensure thorough testing. The results demonstrate that LoD enables the features extracted by SFDA to align more effectively with facial-related information, while better preserving the classification information extracted by LFAD and SFDA networks, without leading to performance degradation. Ultimately, this enhances detection accuracy.
\begin{table}[t]
\caption{Comparison of detection performance with and without LoD across CDFv1, CDFv2, and UADFV datasets. {All metrics in the table are presented as percentages.}
    }
    \centering
    
    \begin{tabular}{c |cc| cc |cc} 
      \toprule  
      \multirow{2}*{\enspace \textbf{LoD } }  &  \multicolumn{2}{c}{\enspace \textbf{CDFv1 } } &  \multicolumn{2}{c}{\enspace \textbf{CDFv2 } } &  \multicolumn{2}{c}{\enspace \textbf{UADFV } } \\
      \cmidrule(lr){2-7}
      &\cellcolor{gray!20} w/  &w/o& \cellcolor{gray!20}w/  &w/o&\cellcolor{gray!20} w/  &w/o \\
      \midrule
      AUC $\uparrow$&   \cellcolor{gray!20}81.22    &77.15 &\cellcolor{gray!20}72.43&72.41&\cellcolor{gray!20}73.63& 71.47
\\
      ACC$\uparrow$ & \cellcolor{gray!20}71.25  & 64.28   &\cellcolor{gray!20}70.68   &68.34&\cellcolor{gray!20}63.51& 60.65
\\
      AP$\uparrow$ &  \cellcolor{gray!20}86.93   & 76.60  &\cellcolor{gray!20}82.95&75.81&\cellcolor{gray!20}74.85&73.12 
\\
      EER $\downarrow$&   \cellcolor{gray!20}24.85   & 34.24 &\cellcolor{gray!20}26.62&37.22&\cellcolor{gray!20}27.85& 26.87
\\
      \bottomrule 
    \end{tabular}
    \label{table:LoD}
\end{table}

\begin{table}[t]
\caption{Performance comparison of DIH networks across datasets using AUC and EER Metrics. {All metrics in the table are presented as percentages.}
    }
    \centering
    \begin{tabular}{c| cc|cc|cc} 
      \toprule  
      \multirow{1}*{\enspace \textbf{DIH} }  &\multicolumn{2}{c} {\textbf{CDFv1}}&\multicolumn{2}{c} {\textbf{CDFv2}} &\multicolumn{2}{c} {\textbf{UADFV}}  \\
      \cmidrule(lr){2-7}
      \enspace \textbf{Network } & AUC $\uparrow$ &EER $\downarrow$& AUC $\uparrow$ &EER $\downarrow$& AUC $\uparrow$ &EER $\downarrow$ \\
      \midrule
      HiNet~\cite{jing2021hinet} &    75.88 & 30.34 &63.22 & 40.58&  75.49   & 25.15
\\
      BalujaNet~\cite{baluja2019hiding} &  66.61  &38.07 & 66.92&38.04&  69.54   & 31.67
\\
      WengNet~\cite{weng2019high} &   72.56  &34.49 &66.19 &38.96 & 65.00    &37.23 
\\
      \cellcolor{gray!20}HiDDeN~\cite{zhu2018hidden} &   \cellcolor{gray!20}81.22  &\cellcolor{gray!20}24.85 &\cellcolor{gray!20}72.43 &  \cellcolor{gray!20}26.62&\cellcolor{gray!20}73.63&\cellcolor{gray!20}27.85
\\
      \bottomrule 
    \end{tabular}
    \label{table:DIH}
\end{table}

Additionally, to demonstrate that the HiDDeN+StegaFFD combination is optimal, we compared the performance of different DIH networks in Table~\ref{table:DIH}. The results show that the HiDDeN+StegaFFD combination outperforms others on most datasets.

\section{Conclusion}
\noindent In this paper, we introduced StegaFFD, a novel privacy-preserving framework for Face Forgery Detection (FFD) that leverages fine-grained steganographic domain analysis. By embedding facial images within natural cover images, StegaFFD ensures covert transmission and analysis, significantly reducing the risk of arousing suspicion from attackers compared to existing privacy-enhancing methods. Our proposed Low-Frequency-Aware Decomposition (LFAD) and Spatial-Frequency Differential Attention (SFDA) modules enable effective feature lifting in the steganographic domain, mitigating interference from cover image semantics. Additionally, the Steganographic Domain Alignment (SDA) enhances detection accuracy by aligning steganographic features with facial information during training. Extensive experiments across seven facial datasets demonstrate that StegaFFD achieves state-of-the-art FFD performance and superior imperceptibility, with only a marginal performance gap compared to non-privacy-preserving methods. However, limitations remain, including slight visual artifacts in stego images and minor performance degradation due to steganographic distortions. Future work will focus on improving imperceptibility by eliminating artifacts and exploring near-lossless FFD frameworks to further bridge the performance gap, enhancing the practicality and robustness of privacy-preserving face forgery detection. StegaFFD paves the way for secure, covert facial analysis in real-world applications.

\section*{Acknowledgments}
This work was supported in part by the National Natural Science Foundation of China (Nos.~62202027 and 62406320), the Strategic Priority Research Program of the Chinese Academy of Sciences (CAS)(XDB1010302), CAS Project for Young Scientists in Basic Research, Grant No. YSBR-041 and the International Partnership Program of the Chinese Academy of Sciences (CAS) (173211KYSB20200021).

\section*{Declarations of Conflict of Interest}

The authors declared that they have no conflicts of interest to this work.

\noindent

\bibliography{sample-base}

\end{document}